\documentclass{capstone}
\usepackage{algorithm}
\usepackage{algpseudocode}
\usepackage{graphicx}
\usepackage{subcaption}

\begin{document}

\title{Data-Efficient American Sign Language Recognition via Few-Shot Prototypical Networks}

\capstoneyear{2025}
\capstonedocument{Seminar}
\capstonesemester{Fall}

\author{Meher Md Saad}
\affiliation{%
  \institution{Computer Science, NYUAD}
}
\email{mehermdsaad@nyu.edu}




\renewcommand{\shortauthors}{Saad}

\begin{abstract}
Isolated Sign Language Recognition (ISLR) is critical for bridging the communication gap between the Deaf and Hard-of-Hearing (DHH) community and the hearing world. However, robust ISLR is fundamentally constrained by data scarcity and the long-tail distribution of sign vocabulary, where gathering sufficient examples for thousands of unique signs is prohibitively expensive. Standard classification approaches struggle under these conditions, often overfitting to frequent classes while failing to generalize to rare ones. To address this bottleneck, we propose a Few-Shot Prototypical Network framework adapted for a skeleton based encoder. Unlike traditional classifiers that learn fixed decision boundaries, our approach utilizes episodic training to learn a semantic metric space where signs are classified based on their proximity to dynamic class prototypes. We integrate a Spatiotemporal Graph Convolutional Network (ST-GCN) with a novel Multi-Scale Temporal Aggregation (MSTA) module to capture both rapid and fluid motion dynamics. Experimental results on the WLASL dataset demonstrate the superiority of this metric learning paradigm: our model achieves 43.75\% Top-1 and 77.10\% Top-5 accuracy on the test set. Crucially, this outperforms a standard classification baseline sharing the identical backbone architecture by over 13\%, proving that the prototypical training strategy effectively outperforms in a data scarce situation where standard classification fails. Furthermore, the model exhibits strong zero-shot generalization, achieving nearly 30\% accuracy on the unseen SignASL dataset without fine-tuning, offering a scalable pathway for recognizing extensive sign vocabularies with limited data.
\end{abstract}


\maketitle

\section{Introduction}

Sign language serves as the primary mode of communication for millions of people with disabling hearing loss worldwide \cite{WHO_Hearing}. For the Deaf and Hard-of-Hearing (DHH) community, sign language is a full-fledged linguistic system with complex grammar, syntax, and vocabulary \cite{stokoe1960sign}. Despite its prevalence, a significant communication gap persists between DHH individuals and the hearing majority who largely lack sign language proficiency. This barrier limits access to essential services, education, and social interaction. Automated Sign Language Recognition (SLR) aims to bridge this gap by translating sign gestures into spoken or written language using advanced machine learning techniques. A fundamental sub-task within this domain is Isolated Sign Language Recognition (ISLR), which focuses on identifying individual distinct words (glosses) from video segments and serves as the foundational building block for more complex continuous sign language translation systems \cite{rastgoo2021sign}.

\begin{figure}[t]
    \centering
    \includegraphics[width=0.95\linewidth]{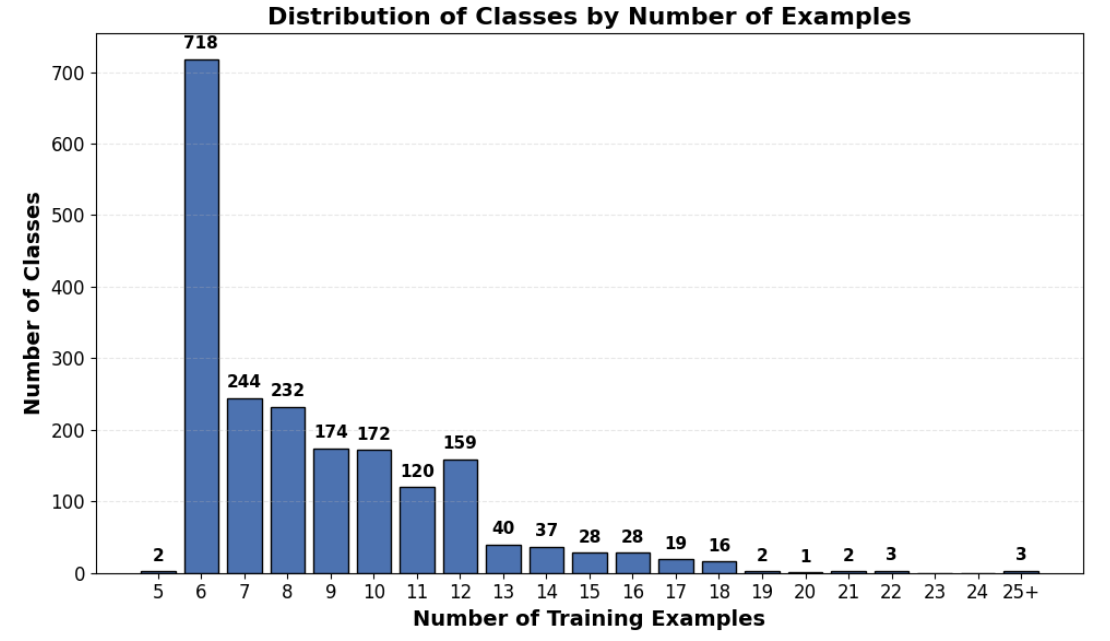}
    \caption{The long-tail class distribution of the WLASL dataset. A small subset of classes contains frequent examples, while the vast majority of signs contain fewer than 13 instances, creating a significant bottleneck for standard classification models.}
    \label{fig:wlasl_dist}
\end{figure}

While deep learning has revolutionized computer vision tasks like object detection, its success is heavily dependent on the availability of massive, balanced datasets. Sign language recognition faces a unique and severe data bottleneck; curating labeled video data for ISLR is more costly than image annotation, requiring native signers, controlled environments, and rigorous temporal segmentation. Consequently, existing benchmarks like the World Level American Sign Language (WLASL) dataset \cite{li2020wlasl} exhibit a severe ``long-tail'' distribution, as illustrated in Figure \ref{fig:wlasl_dist}. A small fraction of common signs have relatively more training examples, while the vast majority of the vocabulary consists of rare signs with as few as five examples. Traditional deep learning classification models, which rely on learning fixed weights for a pre-defined set of classes, struggle in this domain. They tend to overfit to the commonly appearing classes while failing to generalize to the rare signs, making them impractical for real-world applications that require broad vocabulary coverage.

To overcome the limitations of data scarcity, this research explores a \textit{Few-Shot Learning} (FSL) paradigm, specifically utilizing Prototypical Networks \cite{snell2017prototypical}. Instead of training a classifier to memorize specific class boundaries, the model is trained ``episodically'' to learn a generalizable metric space. In this space, an embedding of a query video is classified by finding the nearest ``prototype''—the mean embedding of a small support set of examples. By utilizing a skeleton-based Spatiotemporal Graph Convolutional Network (ST-GCN) \cite{yan2018stgcn} as the backbone encoder, this project focuses on representations that are computationally efficient and robust to background variations.

This seminar report presents the following contributions:
\begin{itemize}
    \item A robust implementation of a Prototypical Network adapted for temporal skeleton-based sign language data, integrating a novel \textbf{Multi-Scale Temporal Aggregation (MSTA)} module to handle varying gesture speeds.
    \item A comparative analysis demonstrating that this few-shot episodic paradigm significantly outperforms standard classifiers on the WLASL dataset (improving Top-1 accuracy by over 13\%).
    \item Evidence of the encoder's strong generalization capabilities, demonstrating effective zero-shot transferability to the unseen SignASL dataset \cite{SignASL} without any fine-tuning.
\end{itemize}

\section{Related Work}

The development of the proposed framework builds on three distinct yet intersecting domains of computer vision research: Isolated Sign Language Recognition (ISLR), Skeleton-based Representation Learning, and Few-Shot Learning (FSL). This section reviews the seminal works in these areas and situates the proposed methodology within the current state of the art.

\subsection{Isolated Sign Language Recognition}
Early approaches to Sign Language Recognition relied heavily on hand-crafted features and Hidden Markov Models (HMMs) to model the temporal dynamics of gestures. With the advent of deep learning, the paradigm shifted towards Convolutional Neural Networks (CNNs). Specifically, 3D-CNNs, such as the I3D architecture proposed by Carreira and Zisserman \cite{carreira2017i3d}, became the standard for capturing spatiotemporal features from full-frame RGB video.

In the specific context of American Sign Language, Li et al. introduced the WLASL dataset \cite{li2020wlasl}, which remains the largest word-level benchmark. They established baselines using 3D-CNNs (I3D) and 2D-CNN-RNN hybrids. While these RGB-based methods achieve high accuracy, they suffer from high computational costs and sensitivity to background clutter and lighting conditions. Furthermore, they typically require massive amounts of training data per class to converge, making them ill-suited for the long-tail distribution often found in real-world sign language vocabularies.

\subsection{Skeleton-Based Representation Learning}
To address the computational inefficiencies and environmental sensitivities of RGB models, recent research has shifted towards skeleton-based recognition. This approach abstracts the human body into a set of keypoints (joints), significantly reducing the data dimension while preserving the semantic information of the movement.

The seminal work by Yan et al. introduced the Spatiotemporal Graph Convolutional Network (ST-GCN) \cite{yan2018stgcn}. Unlike grid-based CNNs, ST-GCNs model the human skeleton as a graph where nodes represent joints and edges represent both natural bone connections and temporal trajectories. This architecture has become the backbone for modern skeleton-based action recognition. Subsequent works have adapted this specifically for sign language; for instance, Jiang et al. \cite{jiang2021skeleton} proposed a Skeleton Aware Multi-modal framework (SAM-SLR) that leverages ST-GCNs to capture fine-grained hand dynamics alongside global body movements.

More recently, Li et al. proposed Uni-Sign \cite{li2024unisign}, a unified framework that achieves state-of-the-art results by pre-training ST-GCN encoders on massive external datasets (e.g., CSL-News) before fine-tuning. While effective, such approaches rely heavily on the availability of large-scale pre-training data to build robust features. In contrast, this research addresses the scenario where such massive external data is unavailable. By utilizing the ST-GCN based encoder as a fixed backbone, this work isolates the impact of the training paradigm, demonstrating that a few-shot episodic learning objective significantly outperforms standard classification when training from scratch on limited samples.

\subsection{Few-Shot Learning and Prototypical Networks}
The challenge of recognizing classes with limited data is addressed by Few-Shot Learning (FSL). Metric-based FSL approaches aim to learn a semantic embedding space where similar samples are clustered together, rather than optimizing a fixed decision boundary.

Snell et al. proposed Prototypical Networks \cite{snell2017prototypical}, a simple yet powerful metric-learning framework. The core idea is to compute a "prototype" vector for each class, defined as the mean of the support set embeddings. Classification is then performed by calculating the Euclidean distance between a query sample and the prototypes. This approach is particularly well-suited for ISLR because signs are defined by "standard" motions; a class prototype ideally represents the canonical version of a sign, averaging out the stylistic variations of individual signers. This research adapts the Prototypical Network strategy to the temporal domain of ST-GCN embeddings, enabling the system to generalize to new signs without extensive retraining.

\section{Methodology}

\subsection{Overview}
We propose a Few-Shot Prototypical Network adapted for the temporal domain of Skeleton-Based Sign Language Recognition (SLR). Unlike traditional methods that require extensive retraining for new vocabulary, our approach learns a semantic metric space where sign trajectories are clustered by similarity. This allows the system to generalize to new signs using only a handful of examples ($K$-shot).

\subsection{Data Preprocessing \& Representation}
To extract robust skeletal features from raw video, we employ a high-performance pose estimation pipeline.

\subsubsection{Pose Extraction (RTMLib)}
We utilize \textbf{RTMLib} \cite{jiang2023rtmpose}, a high-efficiency pose estimation library, to extract 2D keypoints from the input videos. Specifically, we use the \textbf{RTMPose-l} model trained on the \textbf{COCO-WholeBody} dataset\cite{jin2020whole}. This model estimates a dense set of keypoints covering the body, feet, face, and hands, which is crucial for capturing the subtle non-manual markers in sign language.

\subsubsection{Keypoint Partitioning}
We partition the extracted keypoints into a few distinct groups to facilitate the ST-GCN based architecture:
\begin{itemize}
    \item \textbf{Body:} Core joints (shoulders, elbows, wrists, neck) to capture gross motor movements.
    \item \textbf{Hands:} 21 joints per hand to capture fine-grained finger configurations.
    \item \textbf{Face:} A comprehensive set of facial landmarks (including eyebrows, eyes, mouth, and nose) to capture facial expressions and mouthing cues.
\end{itemize}

\subsubsection{Hierarchical Normalization Strategy}
To ensure invariance to camera distance, signer position, and body scale, we apply a three-stage normalization strategy $\mathcal{N}$ that combines global scaling with local relative centering.

\begin{enumerate}
    \item \textbf{Global Body Scaling:} 
    First, we define a global bounding box using the body keypoints. We calculate the scale $s = \max(w, h)$ and center $(c_x, c_y)$ of this box. All keypoints are initially scaled to a normalized range $[-1, 1]$ relative to the body center.
    
    \item \textbf{Local Relative Centering:} 
    To isolate fine-grained motion from global body movement, we apply local subtraction for the hands and face partitions:
    \begin{itemize}
        \item \textbf{Hands:} The position of the wrist joint is subtracted from all other hand keypoints. This centers the hand coordinate system at the wrist ($P'_{hand} = P_{hand} - P_{wrist}$), making the finger configuration invariant to the arm's location in space.
        \item \textbf{Face:} Similarly, the nose tip coordinate is subtracted from all facial points ($P'_{face} = P_{face} - P_{nose}$). This stabilizes the facial expression features against head movements.
    \end{itemize}
    
    \item \textbf{Confidence Gating:} 
    Finally, any keypoints falling below a confidence threshold ($\tau < 0.3$) after normalization are zeroed out to prevent noise propagation from the pose estimator.
\end{enumerate}

\begin{figure*}[t]
    \centering
    \includegraphics[width=\textwidth]{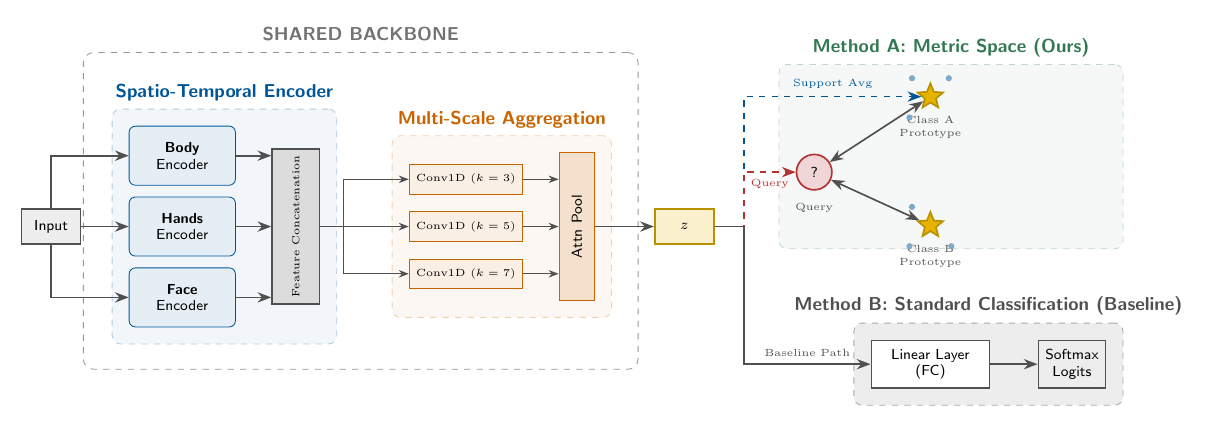}
    \caption{The proposed architecture and experimental comparison. 
\textbf{(Left \& Middle)} The shared backbone: raw skeletal input is partitioned and encoded by an ST-GCN, then aggregated via Multi-Scale Temporal Aggregation (MSTA) into a unified embedding $z$. 
\textbf{(Right-Top) Method A (Ours):} In the Prototypical approach, embeddings map to a metric space where queries are classified by Euclidean distance to class prototypes. 
\textbf{(Right-Bottom) Method B (Baseline):} The baseline approach feeds the same embedding $z$ into a fixed Linear Classification layer. This shared-backbone design ensures fair comparison between the training paradigms.}
    \label{fig:architecture}
\end{figure*}

\subsection{Network Architecture}
Our architecture consists of three distinct modules designed to transform these normalized skeletal sequences into robust embeddings.

\subsubsection{Spatiotemporal Encoder (ST-GCN)}
We utilize a Spatial-Temporal Graph Convolutional Network (ST-GCN) as the backbone encoder $f_{\theta}$. The input consists of the partitions defined above.

\begin{itemize}
    \item \textbf{Spatial Modeling:} Intra-frame dependencies are captured using graph convolutions on the skeletal topology defined by the natural connections of the human body.
    \item \textbf{Fusion:} The partitions are merged via a linear projection to form a unified frame-level embedding sequence $Z \in \mathbb{R}^{T \times D}$, where $D=768$.
    \item \textbf{Structural Inductive Bias:} We leverage the inherent graph structure of the ST-GCN to model geometric dependencies. The network weights are \textbf{initialized randomly} (using a truncated normal distribution) and trained end-to-end. This approach demonstrates that the Prototypical Network can learn robust features directly from the episodic data without relying on external pre-trained weights.
\end{itemize}

\subsubsection{Multi-Scale Temporal Aggregation (MSTA)}
A key addition in our design is the \texttt{MultiScaleTemporalAggregation} module. Sign language gestures vary significantly in speed and duration. A standard global average pooling layer fails to capture fine-grained temporal dynamics.
\begin{itemize}
    \item \textbf{Design Intuition:} We implement parallel 1D Convolutional branches with varying kernel sizes ($k \in \{3, 5, 7\}$).
    \item \textbf{Mechanism:} This allows the network to capture short, sharp movements (small $k$) and longer, fluid gestures (large $k$) simultaneously.
    \item \textbf{Temporal Compression (Attention Pooling):}
    To map the variable-length sequence of frame embeddings $H \in \mathbb{R}^{T \times D}$ into a fixed-size video representation $\mathbf{z} \in \mathbb{R}^{D}$, we employ a learnable attention pooling mechanism.
    
    A scoring network computes a scalar weight $\alpha_t$ for each frame, normalized via Softmax over the temporal dimension. The final video embedding is computed as the weighted sum of all temporal features, effectively compressing the time dimension while preserving the most semantically relevant frames:
    \begin{equation}
        \mathbf{z} = \sum_{t=1}^{T} \text{Softmax}(\alpha_t) \cdot \mathbf{h}_t
    \end{equation}
\end{itemize}

\subsection{Prototypical Training Strategy}
We adapt the Prototypical Network paradigm to the temporal domain. The model is trained in episodes, where each episode is defined by $N$-way classes and $K$-shot support examples.

\subsubsection{Metric Learning Objective}
The goal is to minimize the distance between a query sample and its true class prototype while maximizing the distance to other prototypes.
\begin{itemize}
    \item \textbf{Prototype Computation:} For each class $c$, the prototype $\mathbf{p}_c$ is the mean vector of the embedded support samples $S_c$:
    \begin{equation}
        \mathbf{p}_c = \frac{1}{|S_c|} \sum_{(\mathbf{x}_i, y_i) \in S_c} f_{\theta}(\mathbf{x}_i)
    \end{equation}
    \item \textbf{Classification:} For a query sample $\mathbf{x}_q$, the probability of belonging to class $c$ is based on the softmax over negative Euclidean distances:
    \begin{equation}
        P(y=c|\mathbf{x}_q) = \frac{\exp(-d(f_{\theta}(\mathbf{x}_q), \mathbf{p}_c))}{\sum_{c'} \exp(-d(f_{\theta}(\mathbf{x}_q), \mathbf{p}_{c'}))}
    \end{equation}
\end{itemize}

\begin{algorithm}[H]
\caption{Prototypical ST-GCN Training Episode}
\label{alg:training_episode}
\begin{algorithmic}[1]
\Require Support Set $\mathcal{S}$, Query Set $\mathcal{Q}$
\Require ST-GCN Encoder $f_\theta$, Aggregator $A_\phi$
\Ensure Updated parameters $\theta, \phi$

\Procedure{TrainEpisode}{$\mathcal{S}, \mathcal{Q}$}
    \State \textbf{1. Feature Extraction:}
    \State $Z_S \leftarrow A_\phi(f_\theta(\mathcal{S}))$ \Comment{Encode \& Aggregate Support}
    \State $Z_Q \leftarrow A_\phi(f_\theta(\mathcal{Q}))$ \Comment{Encode \& Aggregate Query}

    \State \textbf{2. Compute Prototypes:}
    \State $\mathbf{p}_c \leftarrow \frac{1}{|S_c|} \sum_{(\mathbf{x},y) \in \mathcal{S}_c} Z_{S}^{(\mathbf{x})}$ \quad $\forall c \in \{1 \dots N\}$
    \Comment{Class Centroids}

    \State \textbf{3. Metric Classification:}
    \State $\mathcal{D} \leftarrow \text{EuclideanDist}(Z_Q, \{\mathbf{p}_c\})$ \Comment{Compute distances}
    \State $\hat{Y} \leftarrow \text{Softmax}(-\mathcal{D})$ \Comment{Convert distances to probabilities}

    \State \textbf{4. Optimization:}
    \State $J \leftarrow \text{CrossEntropyLoss}(\hat{Y}, Y_{true})$
    \State Update $\theta, \phi$ using $\nabla J$ with Mixed Precision Scaling

    \State \Return $J$
\EndProcedure
\end{algorithmic}
\end{algorithm}

\subsection{Training Adaptations}
We engineered several domain-specific optimizations in our training pipeline to make it suitable for our use-case:

\begin{enumerate}
    \item \textbf{Global Prototype Validation (Open-Set Simulation):}
    Standard few-shot papers validate on small, isolated episodes (e.g., 5-way). However, real-world SLR is an open-set problem. Our validation strategy constructs a \textit{Global Prototype Dictionary} using the entire training set. We test the model's ability to distinguish a new sign against \textit{all} known signs, ensuring the learned embeddings are globally separable rather than just locally distinct.
    
    \item \textbf{Automatic Mixed Precision (AMP) \& Collated Sampling:}
    Processing pose sequences is memory-intensive for higher N way settings. We implemented an Automatic Mixed Precision pipeline (via \texttt{torch.cuda.amp}) combined with a custom \texttt{EpisodicBatchSampler}. This engineering optimization allows us to scale the episode size ($N$-way) significantly higher than standard baselines. Training with higher $N$-way is mathematically shown to produce more robust embeddings as the model must distinguish the target from a larger set of distractors.
    
    \item \textbf{Temporal Speed Augmentation:}
    We apply random speed scaling factors ($0.8\times, 1.0\times, 1.25\times$) during data loading. This forces the model to learn the \textit{relative} motion patterns (trajectory shape) rather than memorizing the absolute duration of specific video files.
\end{enumerate}

\section{Experiments}

\subsection{Experimental Setup}
\subsubsection{Datasets}
\textbf{WLASL:} We trained our primary models on the WLASL dataset, a large-scale word-level sign language recognition benchmark containing over 2,000 classes.

\textbf{SignASL (Evaluation Only):} In addition to WLASL, we evaluate our model on the SignASL.com dataset \cite{SignASL}. While this dataset is primarily accessible for online browsing, our research laboratory obtained bulk access for research purposes. In this project, SignASL was used \textbf{exclusively for evaluation}. We intentionally excluded it from the training set to measure cross-dataset generalization to data collected under different recording conditions and signer demographics.

\subsubsection{Implementation Details}
\begin{itemize}
    \item \textbf{Backbone:} We employed an ST-GCN encoder combined with our proposed Multi-Scale Temporal Aggregator (MSTA). The encoder was initialized with random weights (trained from scratch) to validate the model's ability to learn representations directly from episodic data.
    \item \textbf{Training Protocol:} Models were trained for 50,000 episodes with $K=3$ shots and $Q=2$ query samples per class using the Adam optimizer.
    \item \textbf{Baselines:} We compare against a standard classification baseline sharing the identical ST-GCN + MSTA backbone but utilizing a fully connected linear classifier trained for 50 epochs.
\end{itemize}

\subsection{Quantitative Results}

\subsubsection{\textbf{Comparison with Standard Classification Baseline}}
Our primary evaluation compares the Prototypical ST-GCN against the standard classification baseline on the WLASL test set. We report Top-1, Top-5, and Top-10 accuracy to account for the semantic ambiguity often present in sign language recognition.

As shown in Table \ref{tab:main_results}, our metric learning approach significantly outperforms the traditional classification head across all metrics.

\begin{table}[h]
\centering
\caption{Performance comparison on WLASL (Test Set). The Prototypical approach (trained with 100-Way episodes) outperforms the standard classification baseline by over 13\% in Top-1 accuracy and nearly 18\% in Top-5 accuracy.}
\label{tab:main_results}
\resizebox{\columnwidth}{!}{%
\begin{tabular}{llccc}
\hline
\textbf{Method} & \textbf{Configuration} & \textbf{Top-1} & \textbf{Top-5} & \textbf{Top-10} \\ \hline
Baseline & Standard Classifier (50 Epochs) & 30.37\% & 59.24\% & 70.26\% \\
\textbf{Ours} & \textbf{Prototypical (200-Way)} & \textbf{43.75\%} & \textbf{77.10\%} & \textbf{84.50\%} \\ \hline
\end{tabular}%
}
\end{table}

The baseline achieves only 30.37\% Top-1 accuracy. In contrast, our best-performing configuration (200-Way) achieves 43.75\%. Notably, the Top-5 accuracy gap is even wider ($17.86\%$), suggesting that while the baseline frequently misses the correct sign entirely, our model usually places the correct sign within the top candidates, making it far more practical for real-world assistive recommendation systems.

\subsubsection{\textbf{Ablation Study: Impact of N-Way Training}}
We investigated the impact of the ``$N$-way'' hyperparameter (the number of classes per training episode) on model performance. We hypothesized that higher $N$-way training acts as a harder negative mining strategy, forcing the model to learn more discriminative embeddings.

Table \ref{tab:nway_ablation} summarizes the results on the WLASL Test and Validation sets.

\begin{table}[h]
\centering
\caption{Impact of Episode Size ($N$-way) on model performance. Increasing $N$ yields consistent gains across all metrics, with the 200-Way configuration achieving the highest generalization performance.}
\label{tab:nway_ablation}
\resizebox{\columnwidth}{!}{%
\begin{tabular}{lcccc}
\hline
\textbf{N-Way} & \textbf{Val Acc} & \textbf{Test Top-1} & \textbf{Test Top-5} & \textbf{Test Top-10} \\ \hline
50-Way & 42.98\% & 40.62\% & 75.40\% & 83.46\% \\
100-Way & 44.79\% & 43.43\% & 76.79\% & 84.02\% \\
\textbf{200-Way} & \textbf{45.56\%} & \textbf{43.75\%} & \textbf{77.10\%} & \textbf{84.50\%} \\ \hline
\end{tabular}%
}
\end{table}

The results confirm that increasing the difficulty of the training episodes improves generalization. The model trained with 100-Way episodes outperforms the 50-Way model by nearly 3\% in Top-1 accuracy. Furthermore, scaling to \textbf{200-Way} yields the best overall performance, achieving \textbf{43.75\%} Top-1 and \textbf{84.5\%} Top-10 accuracy on the test set. This strictly positive trend validates our hypothesis: training with a larger number of classes per episode forces the encoder to learn sharper decision boundaries to distinguish the correct prototype from a larger set of negatives, resulting in more robust embeddings that generalize better to unseen data.
\subsubsection{\textbf{Cross-Dataset Generalization (SignASL)}}
To assess truly out-of-distribution robustness, we evaluated the WLASL-trained models directly on the \textbf{SignASL} dataset (19,000 videos). 
\textbf{Note:} This evaluation is zero-shot domain transfer. The prototypes (support set) were constructed using WLASL data, while the queries (test samples) came from SignASL. This forces the model to match unseen videos from a different domain to its learned WLASL representations.

\begin{table}[h]
\centering
\caption{Zero-Shot Cross-Dataset Evaluation on SignASL. The Prototypical model demonstrates superior domain transfer compared to the fixed-classifier baseline.}
\label{tab:signasl_results}
\resizebox{\columnwidth}{!}{%
\begin{tabular}{lccc}
\hline
\textbf{Method} & \textbf{Top-1} & \textbf{Top-5} & \textbf{Top-10} \\ \hline
Baseline (Standard Classifier) & 24.12\% & 37.76\% & 42.83\% \\
\textbf{Ours (Prototypical)} & \textbf{29.87\%} & \textbf{44.00\%} & \textbf{48.24\%} \\ \hline
\end{tabular}%
}
\end{table}

Despite the significant domain shift (different signers, lighting, and recording pipelines), our Prototypical model maintains a Top-1 accuracy of nearly 30\%, outperforming the baseline by over 5\%. This indicates that the metric space learned by our ST-GCN encoder captures generic kinematic signatures of sign language that persist across different datasets. In contrast, the \textbf{Standard Classification baseline}- which relies on fixed weight matrices learned specifically for WLASL classes- overfits more heavily to the specific visual artifacts of the source domain and fails to generalize as effectively to the new distribution.
\subsection{Qualitative Analysis}
To better understand the internal representations learned by the model, we conducted a qualitative inspection of the latent space and an error analysis.

\subsubsection{\textbf{Latent Space Visualization}}
We projected the high-dimensional embeddings ($D=768$) into 2D space using Principal Component Analysis (PCA).

\begin{figure*}[t]
    \centering
    \includegraphics[width=\textwidth]{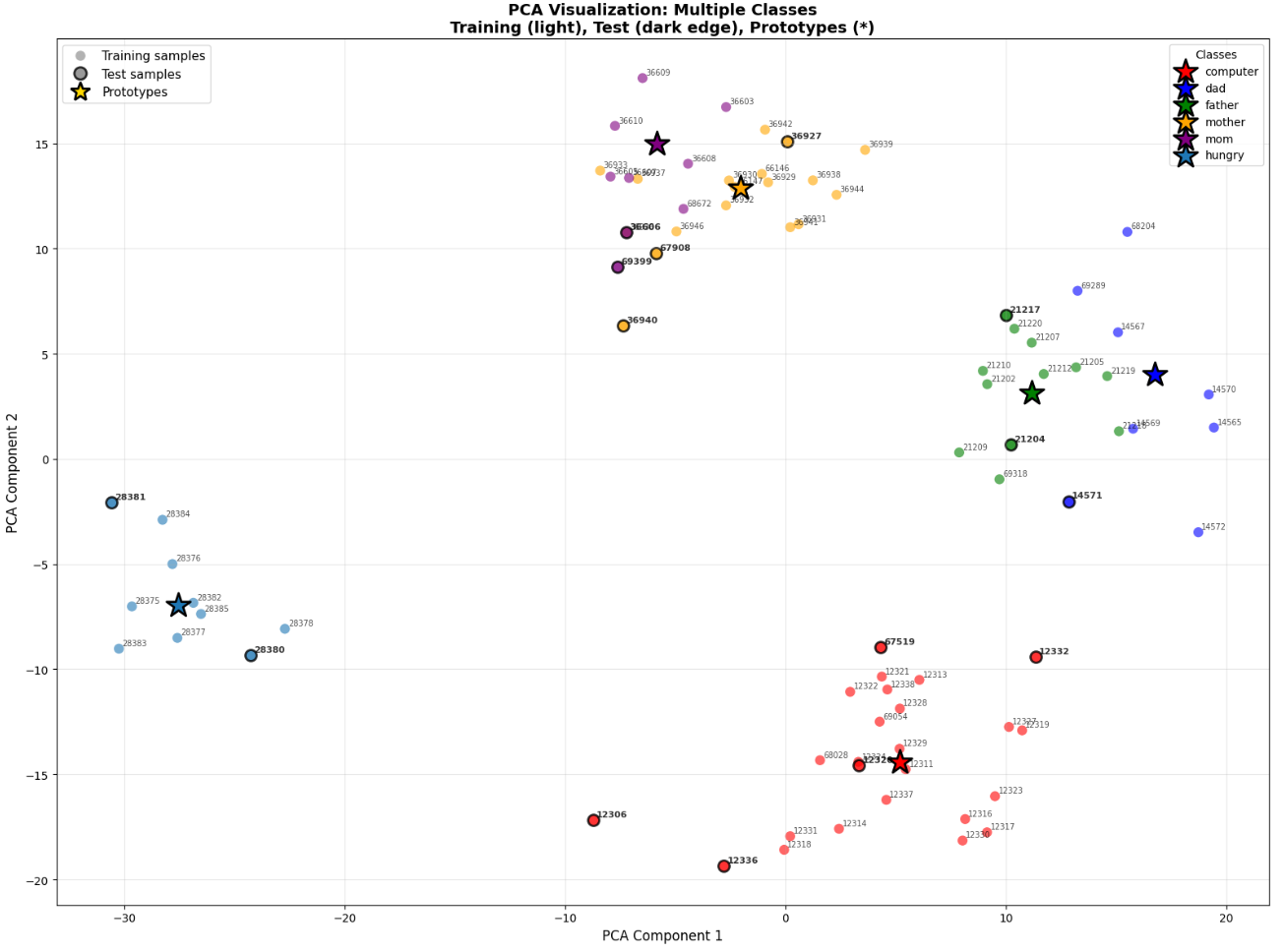}
    \caption{PCA projection of the latent space for selected classes. Semantically unrelated signs (e.g., ``computer'', ``hungry'') form distinct, isolated clusters. In contrast, synonymous pairs (``mom''/``mother'' and ``dad''/``father'') form overlapping clusters, demonstrating that the metric space captures semantic equivalence through kinematic similarity.}
    \label{fig:qualitative}
\end{figure*}

\textbf{Observation:} As illustrated in Figure \ref{fig:qualitative}, the learned embedding space exhibits strong structural organization. We observe that semantically unrelated concepts such as ``computer'' (red) and ``hungry'' (light blue) are mapped to distant, well-separated regions. Conversely, synonymous terms naturally cluster together without explicit semantic supervision. The classes ``mom'' and ``mother'' (top center) and ``dad'' and ``father'' (right) form tight, overlapping groups. This confirms that the model successfully infers semantic proximity purely from the kinematic similarity of the pose sequences.

\subsubsection{\textbf{Error Analysis: Synonym vs. Kinematic Ambiguity}}
To quantify the nature of the model's errors, we extracted the top confusing pairs from the validation set. Table \ref{tab:confused_pairs} lists the 15 class pairs with the highest mutual confusion rates.

\begin{table}[h]
\centering
\caption{Top 15 most confused class pairs from WLASL test data. A significant majority of these pairs are linguistic synonyms (e.g., \textit{jail/prison}, \textit{phone/telephone}), confirming that the model confuses labels rather than concepts.}
\label{tab:confused_pairs}
\resizebox{0.9\columnwidth}{!}{%
\begin{tabular}{lllc}
\hline
\textbf{Rank} & \textbf{Class A} & \textbf{Class B} & \textbf{Count} \\ \hline
1 & desk & table & 5 \\
2 & angel & fairy & 3 \\
3 & breeze & wind & 3 \\
4 & build & building & 3 \\
5 & california & gold & 3 \\
6 & depend & rely & 3 \\
7 & great & wonderful & 3 \\
8 & hug & love & 3 \\
9 & jail & prison & 3 \\
10 & phone & telephone & 3 \\
11 & poor & poverty & 3 \\
12 & soft & wet & 3 \\
13 & tea & vote & 3 \\
14 & temperature & thermometer & 3 \\
15 & accident & crash & 2 \\ \hline
\end{tabular}%
}
\end{table}

\begin{figure}[h]
    \centering
    \begin{subfigure}[b]{0.48\columnwidth}
        \centering
        \includegraphics[width=\textwidth]{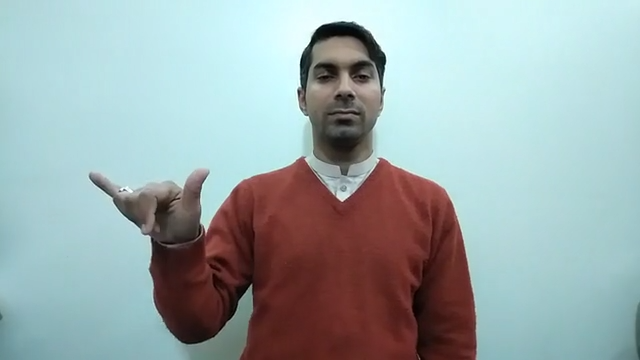}
        \caption{Sign for ``Gold''}
        \label{fig:gold}
    \end{subfigure}
    \hfill 
    \begin{subfigure}[b]{0.48\columnwidth}
        \centering
        \includegraphics[width=\textwidth]{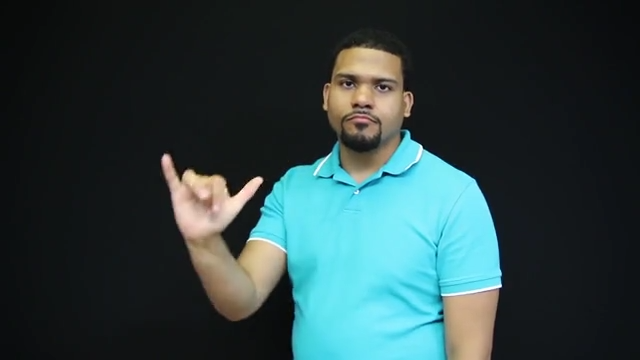}
        \caption{Sign for ``California''}
        \label{fig:california}
    \end{subfigure}
    
    \caption{Visual comparison of kinematic homophenes. Despite having distinct meanings, the signs for (a) ``Gold'' and (b) ``California'' share nearly identical skeletal configurations and motion trajectories, leading to inherent ambiguity in the metric space.}
    \label{fig:gold_vs_california}
\end{figure}

\begin{figure}[h]
    \centering
    \begin{subfigure}[b]{0.48\columnwidth}
        \centering
        \includegraphics[width=\textwidth]{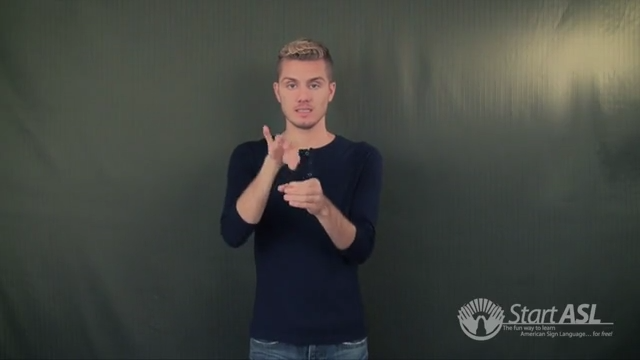}
        \caption{Sign for ``Tea''}
        \label{fig:tea}
    \end{subfigure}
    \hfill 
    \begin{subfigure}[b]{0.48\columnwidth}
        \centering
        \includegraphics[width=\textwidth]{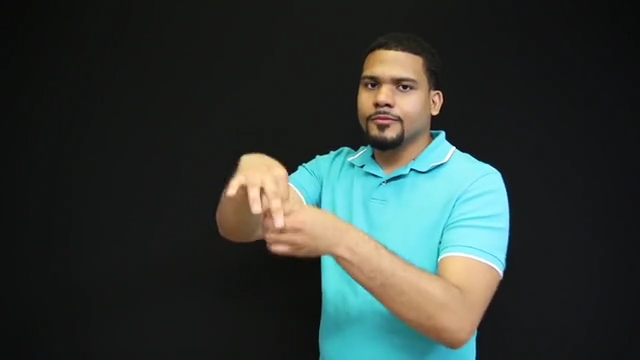}
        \caption{Sign for ``Vote''}
        \label{fig:vote}
    \end{subfigure}
    
    \caption{Visual comparison of kinematic homophenes. The signs for (a) ``Tea'' (stirring motion) and (b) ``Vote'' (insertion motion) are semantically unrelated but share a highly similar skeletal trajectory and hand placement, leading to confusion in the ST-GCN embedding space.}
    \label{fig:tea_vs_vote}
\end{figure}

The data in Table \ref{tab:confused_pairs} strongly supports our hypothesis regarding semantic and kinematic similarity.

\begin{enumerate}
   \item \textbf{Synonym Ambiguity (Labeling Granularity):} 
The majority of the top confusions correspond to direct synonyms or grammatical variations of the same root concept (e.g., \textit{Build/Building}, \textit{Poor/Poverty}, \textit{Accident/Crash}). In many sign language annotation schemas, these terms map to the exact same physical sign execution. The model effectively learns the correct visual-semantic concept but is penalized for predicting a synonymous label that the dataset treats as a distinct class. This suggests that the model's "errors" in these cases are often linguistically valid interpretations, exposing the limitations of rigid class boundaries in the dataset rather than a failure of the recognition system.
    
    \item \textbf{Kinematic Ambiguity (Historical \& Visual):} Some pairs are distinct concepts that share a high degree of visual similarity. A prime example is \textbf{\textit{California/Gold}} (Rank 5). As illustrated in Figure \ref{fig:gold_vs_california}, the skeletal configuration for these two signs is nearly indistinguishable in individual frames. This is rooted in ASL etymology, where the sign for ``California'' is derived from the sign for ``Gold'' (referencing the Gold Rush). Because our ST-GCN encoder relies on skeletal pose data, it correctly maps these visually identical motion patterns to overlapping regions in the embedding space, resulting in unavoidable confusion.
\end{enumerate}

\section{Conclusion \& Future Work}

\subsection{Conclusion}
In this work, we proposed a Few-Shot Prototypical Network adapted for the temporal domain of Skeleton-Based Sign Language Recognition. By integrating a decoupled ST-GCN encoder with a Multi-Scale Temporal Aggregation (MSTA) module, our framework successfully learns a robust metric space where sign trajectories are organized by semantic similarity rather than just visual patterns.

Our extensive experiments on the WLASL dataset demonstrate that this metric learning paradigm is significantly more effective for long-tail distributions than standard classification approaches. Our approach achieved a \textbf{43.75\% Top-1 accuracy} on the test set, outperforming the standard classification baseline by over \textbf{13\%}, and demonstrated superior generalization with an \textbf{18\% gain in Top-5 accuracy}. Furthermore, the model exhibited strong zero-shot transfer capabilities, achieving nearly \textbf{30\% accuracy} on the unseen SignASL dataset without any fine-tuning.

Qualitative analysis of the latent space revealed a key insight: our model spontaneously discovers semantic structures, clustering synonymous terms like ``Mom/Mother'' and ``Dad/Father'' together. This suggests that the ``errors'' often reported in SLR metrics Sare frequently valid linguistic interpretations penalized by rigid dataset labeling.

\subsection{Future Work}
Building on these findings, we identify three promising directions for future research:

\begin{enumerate}
    \item \textbf{Semantic Label Refinement:} 
    Our error analysis highlighted that a significant portion of misclassifications arises from synonymous labels (e.g., \textit{jail/prison}). Future work should explore \textbf{Soft Labeling} or \textbf{similarity-aware evaluation metrics} that penalize synonym confusion less severely than unrelated errors. This would provide a fairer assessment of a model's communicative utility in real-world scenarios.

    \item \textbf{Expert-Based Non-Manual Analysis:} 
    While our current ST-GCN based encoder does include facial keypoints, relying solely on raw geometric coordinates may be insufficient for distinguishing minimal pairs (e.g., \textit{Tea/Vote}) that depend on subtle mouthing or expression cues. Future iterations could integrate \textbf{expert-based sub-modules} (specifically pre-trained for lip-reading or facial expression recognition) to extract higher-level semantic features from the face, rather than treating it purely as a skeletal graph partition.

    \item \textbf{Adaptive \& Dynamic Prototyping:} 
    Currently, we represent each class as a fixed point (centroid). However, complex signs often have higher variance (larger cluster spread) than simple signs. Future iterations could employ \textbf{probabilistic embeddings} (modeling prototypes as distributions with learnable variance) to handle these differing cluster densities, or use attention-based mechanisms to dynamically refine prototypes based on the query context.

    \item \textbf{Self-Supervised Pre-training:} 
    While our current model relies on random initialization, we believe performance could be further boosted by exploring \textbf{self-supervised or weakly-supervised pre-training} strategies (e.g., masked pose modeling). Pre-training the encoder on vast amounts of unlabeled sign language video could establish stronger kinematic priors before fine-tuning with our few-shot prototypical paradigm.
\end{enumerate}

\bibliographystyle{ACM-Reference-Format}
\bibliography{refs}

\end{document}